\documentclass[sigconf, nonacm]{acmart}

\usepackage{subfiles}
\usepackage{booktabs} 
\usepackage{caption}
\usepackage{subcaption}
\usepackage{graphicx}
\usepackage{multirow}
\usepackage{enumitem}
\usepackage{balance}

\newcommand\vldbdoi{XX.XX/XXX.XX}
\newcommand\vldbpages{XXX-XXX}
\newcommand\vldbvolume{15}
\newcommand\vldbissue{3}
\newcommand\vldbyear{2022}

\newcommand\vldbtitle{\shorttitle} 
\newcommand\vldbavailabilityurl{URL_TO_YOUR_ARTIFACTS}
\newcommand\vldbpagestyle{empty}

\newcommand{\paratitle}[1]{\vspace{1ex}\noindent \textbf{#1}}

\begin{document}
\title{Points-of-Interest Relationship Inference with Spatial-enriched Graph Neural Networks}

\author{Yile Chen$^{1}$, Xiucheng Li$^{1}$, Gao Cong$^{1}$, Cheng Long$^1$, Zhifeng Bao$^2$, Shang Liu$^1$, \\Wanli Gu$^3$, Fuzheng Zhang$^3$}
\affiliation{
\institution{$^1$Nanyang Technological University \;\;\;\;\ $^2$RMIT University \;\;\;\;\ $^3$Meituan}
\country{}
}
\email{{yile001@e.,xli055@e.,gaocong@, c.long@, shang006@e.}ntu.edu.sg,}
\email{zhifeng.bao@rmit.edu.au, {guwanli, zhangfuzheng}@meituan.com}

\begin{abstract}
    As a fundamental component in location-based services, inferring the relationship between points-of-interests (POIs) is very critical for service providers to offer good user experience to business owners and customers.
    Most of the existing methods for relationship inference are not targeted at POI, thus failing to capture unique spatial characteristics that have huge effects on POI relationships. 
    In this work we propose \textsf{PRIM} to tackle POI relationship inference for multiple relation types. \textsf{PRIM} features four novel components, including a weighted relational graph neural network, category taxonomy integration, a self-attentive spatial context extractor, and a distance-specific scoring function.
    Extensive experiments on two real-world datasets show that \textsf{PRIM} achieves the best results compared to state-of-the-art baselines and 
    it is robust against data sparsity and is applicable to unseen cases in practice.
\end{abstract}

\maketitle

\pagestyle{\vldbpagestyle}
\begingroup\small\noindent\raggedright\textbf{PVLDB Reference Format:}\\
Yile Chen, Xiucheng Li, Gao Cong, Cheng Long, Zhifeng Bao, Shang Liu, Wanli Gu, Fuzheng Zhang.
\vldbtitle. PVLDB, \vldbvolume(\vldbissue): \vldbpages, \vldbyear.\\
\href{https://doi.org/\vldbdoi}{doi:\vldbdoi}
\endgroup
\begingroup
\renewcommand\thefootnote{}\footnote{\noindent
This work is licensed under the Creative Commons BY-NC-ND 4.0 International License. Visit \url{https://creativecommons.org/licenses/by-nc-nd/4.0/} to view a copy of this license. For any use beyond those covered by this license, obtain permission by emailing \href{mailto:info@vldb.org}{info@vldb.org}. Copyright is held by the owner/author(s). Publication rights licensed to the VLDB Endowment. \\
\raggedright Proceedings of the VLDB Endowment, Vol. \vldbvolume, No. \vldbissue\ %
ISSN 2150-8097. \\
\href{https://doi.org/\vldbdoi}{doi:\vldbdoi} \\
}\addtocounter{footnote}{-1}\endgroup


\section{Introduction}\label{sec:intro}

With fast development of urban intelligence, points-of-interest (POI) has been extensively utilized in numerous location-based services (e.g., Yelp, Meituan). Massive POIs enable us to conduct various analysis to facilitate local business ranging from marketing to delivery. POIs are correlated explicitly or implicitly under different relationships, among which competitive and complementary relationships have important applications and have been studied in previous work~\cite{KDD15_itemrelation,WSDM18_itemrelation,CIKM20_DecGCN,WSDM19_itemrelation}. In general, it is of great importance for service providers to have a deep understanding of POIs, as well as their latent relationships. 
Specifically, given POIs and some available relationships among them, such as competitive and complementary relationships, and side information such as category taxonomy, we aim to infer other missing relationships. POI relationship inference can bring significant benefits for different groups of people. For example, business owners can design targeted operation strategies to attract more customers according to competitive POIs, 
and customers can be recommended with places of their interests based on complementary POIs.
Furthermore, the study of relationships among POIs can help government understand  regional functionality in terms of industry, commerce and lifestyle, and hence make sustainable urban plannings.
In practice, POI relationship inference model has been utilized in Meituan to provide an automatic and accurate way of enriching internal spatial knowledge graph, which serves as a type of data source to be applied in various business scenarios, such as search, tagging and  recommendation. For example, it has been leveraged in knowledge-enhanced models~\cite{KGRec_Survey} to improve the performance of POI recommendation~\cite{KDD19}.

A number of studies have been conducted to infer the relationships between entities in various domains, especially for companies in business management and products in e-commerce. Earlier approaches focus on modeling text content (e.g., reviews and news) via topic modeling techniques~\cite{CIKM12_company, KDD15_itemrelation} or constructing pairwise neural networks~\cite{WSDM19_itemrelation}. Recently, graph representation learning methods~\cite{NE_Survey} have been applied to solve this problem by organizing entities as a graph with different model designs, such as adding path constraints~\cite{WSDM18_itemrelation}, multi-task learning objectives~\cite{WWW20_company,KDD20_company} and Graph Neural Networks (GNN)~\cite{CIKM20_DecGCN}. However, these approaches are not primarily designed for POIs and do not explore unique features brought by POIs. Moreover, some of them may not even be applicable to POI relationship inference problem~\cite{KDD20_company,WWW20_company}. 
Meanwhile, there are very few approaches tailored for POI settings. Zhou et al.~\cite{TKDE_POI} propose to extract POI pairwise features by sampling a large number of graphlet patterns, and then send these features to neural networks to infer the relationship. The most recent method, DeepR~\cite{KDD20_POI}, is proposed to adapt GNN to perform neighbor aggregation from different geographical sectors, and incorporate brand and aspect knowledge from external data sources to improve the model performance.       

Although these methods, including POI-specific methods and non POI-specific methods that can be applicable to POIs, have 
made progress in inferring the relationships for POIs, 
they still suffer from at least one of the following issues. 
\underline{Issue 1} -- Some methods only focus on a particular type of relationship. For example, recent POI-specific methods~\cite{TKDE_POI, KDD20_POI} consider  the competitiveness relationship only. While they can be extended to multiple relationships by decomposing relationship graph into multiple sub-graphs with each sub-graph containing only one relation type,
these methods fail to model the inherent interactions of POIs under different types of relationships in a unified framework.
\underline{Issue 2} -- As a unique property of POI, spatial features have important effects on POI relationships. For example, two Starbucks would exhibit a strong \emph{competitive} relationship if they are close, but have no relationship if far away. However, such spatial features are not captured in either representation learning modules~\cite{TKDE_POI} or predictive functions~\cite{KDD20_POI} in previous work,  thus leading to suboptimal model performance. 
\underline{Issue 3} -- Some methods propose to enhance the model with structural knowledge (e.g., fuzzy logics~\cite{WSDM18_itemrelation} or graphlet patterns~\cite{TKDE_POI}). However, they require either heavy processing operations or efforts on manually designed constraints. There is a lack of simple and direct ways to integrate extra structural knowledge into the model. 
\underline{Issue 4} -- All existing methods focus on utilizing the relationships of interest without considering the context information of target POIs. In other words, for a target POI, they only focus on its connected POIs that have relationships with it. Apart from connected POIs, the POIs that are spatially close to the target POI can also provide rich context information. For example, given a POI with shopping centers and entertainment spots nearby, we can infer that it is located at a commercial area, which indicates a different degree of competitive environment as compared to a residential area. Such context information can be useful for relationship inference.

To address these issues, we propose a novel \emph{P}OI \emph{R}elationship \emph{I}nference \emph{M}odel, named \textsf{PRIM}, which can handle multiple relation types within a unified framework. 
\underline{First}, to address Issue 1, we propose to 
build \textsf{PRIM} upon the weighted relational graph neural network (WRGNN), in which we apply a two-level aggregation process, namely intra-relationship aggregation and inter-relationship aggregation, to update POI representations. We model the influence on the target representation aggregated from different relationships with a relation-specific operator that explicitly models interactions between a neighbor POI and its relationship.
\underline{Second}, we propose a spatial-aware attention mechanism for WRGNN to measure the importance of different neighbors by considering the POI spatial characteristics, which alleviates Issue 2 in representation learning module. 
Moreover, we design a distance-specific scoring function that is able to model the impact of pairwise distance of a given POI pair. By using this scoring function, we compute the likelihood for different relationship, which further enhances the model capacity in predictive function for the second issue.
\underline{Third}, we address Issue 3 by utilizing category taxonomy as external structural knowledge to enhance reasoning capacity of the model. Specifically, we extract the category path from the category taxonomy and seamlessly integrate it to guide the reasoning of WRGNN via embedding learning method without requiring any extra feature generation procedures or human efforts. 
\underline{Fourth}, apart from learning based on POIs with relationships, we propose to enrich the POI representations from spatial neighbors with a self-attentive spatial context extractor which combines semantic and geographical influence together, which helps remedy Issue 4.

The main contributions of our work are summarized as follows:
\vspace{-1ex}
\begin{itemize} 
    \item To the best of our knowledge, we are the first to propose a unified framework to handle different relationships for various business scenarios.
    \item We propose a full-fledged solution called  \textsf{PRIM}, which features several novel designed techniques, including a weighted relational graph neural network, category taxonomy integration, a self-attentive spatial context extractor and a distance-specific scoring function. 
    These techniques are designed to address the aforementioned four issues encountered by the existing relationship inference models.
    \item We conduct comprehensive experiments on two real-world datasets to evaluate the effectiveness of the proposed model. Experimental results show that \textsf{PRIM} outperforms the state-of-the-art baselines, including relationship inference models as well as graph representation learning models.
\end{itemize}

\section{Related Work}\label{sec:relatedwork}

 In this section, we present related work on relationship inference, POI mining, and graph neural networks.

\paratitle{Relationship Inference.}
Relationship inference has attracted much attention in various domains for practical usage, especially in business management and e-commerce.
In business management, company relationship is analysed to guide strategic planning. Early work adopts graphical models to mine competitive relationship based on text and social network data~\cite{company11, CIKM12_company}. In recent studies~\cite{KDD20_company,WWW20_company} the authors propose to construct a company relationship graph in different ways and then combine multi-task learning strategies with graph representation learning techniques~\cite{NE_Survey, GNN_Survey} to conduct relationship inference.  
In e-commerce, product relationship inference is crucial to  improving online advertisement and recommendation services. Given the product reviews as input, the authors in~\cite{KDD15_itemrelation} employ topic modeling to generate product topic distribution followed by a classifier to predict the complementary and substitutable relationship between two products. LVA~\cite{WSDM19_itemrelation} is then proposed to further improve the performance by adopting variational autoencoders to model product reviews. Unfortunately, these methods consider each product pair \emph{independently} and ignore rich relationship information from the underlying product graph. 
To handle this issue, some methods are proposed to apply graph representation learning techniques on the product graph for relationship inference by incorporating heuristic path constraints~\cite{WSDM18_itemrelation}, preserving node proximity~\cite{IJCAI19} or adopting Graph Neural Networks~\cite{CIKM20_DecGCN}. 
Some of these methods can be extended for POI relationship inference. However, they suffer from several weaknesses discussed in Section~\ref{sec:intro}. There are very limited studies on POI relationship inference. Zhou et al.~\cite{TKDE_POI} and Li et al.~\cite{KDD20_POI} propose to perform POI competitive analysis, but they have several weaknesses as discussed in Section~\ref{sec:intro}.  

\paratitle{POI Mining.}
POIs have been utilized to facilitate numerous applications in location-based services. Extensive studies have been conducted on POI recommendation based on user historical check-ins to understand user behavior patterns and preferences for POIs~\cite{VLDB_POIrec,SIGIR20_POI}. Furthermore, POI is leveraged in a broader view to enhance urban intelligence~\cite{UrbanComputing}. 
For example, POIs associated with other data sources (e.g., text) contain richer knowledge about  locations and are utilized in urban event detection~\cite{SIGIR_event, VLDBJ_event}, human mobility modeling~\cite{KDD_mobility, WSDM_mobility, WSDM20_DeepJMT}, etc. On the other hand, to better support the above applications, methods are proposed to improve the quality of data management of POIs, especially when combined with other data sources, including streaming scenarios~\cite{VLDB_streamPOI}, visualization~\cite{SIGMOD_POIvis}, scalability~\cite{SIGMOD_POIscalability} and efficient query retrieval~\cite{VLDB_POIretrieve, VLDBJ_query}. 
These studies are orthogonal to our problem of mining relationships among POIs, and POI relationships can be used to improve on these applications, such as POI recommendation.

\paratitle{Graph Neural Networks.}
Graph Neural Networks (GNN)~\cite{GNN_Survey} is a generic method on modeling graph-structured data and has achieved great successes in learning effective node representations~\cite{VLDB19_AliGNN}. Conventional GNN~\cite{GCN, GraphSAGE, GAT} performs message passing and message aggregation from neighbors for each node iteratively to update node representations. However, they are less capable of dealing with heterogeneous graphs consisting of various node-and-edge types which usually exist in real-world applications. To overcome this issue, some methods, namely Heterogeneous Graph Neural Networks (HGNN), have been proposed to adapt GNN to heterogeneous graphs, and they can be roughly divided into two categories. The first is to convert a heterogeneous graph into multiple graphs based on pre-defined meta-paths, and then aggregate information from multiple graphs to obtain final representations~\cite{WWW19_HAN, WWW20_MAGNN, TKDE_HGNN}. 
However, they demand the domain expertise in good meta-paths design, which is both labor-intensive and expensive to acquire.
The second is to modify the conventional message passing and aggregation to be node-type or edge-type dependent operations to integrate various node and edge type information~\cite{WWW20_HGT, RGCN, ICLR20_CompGCN}. Unfortunately, they either ignore different importance of node neighbors, or fail to capture unique spatial characteristics of POIs. The state-of-the-art relationship inference methods, DeepR~\cite{KDD20_POI} and DecGCN~\cite{CIKM20_DecGCN}, fall into conventional GNN and the second type of HGNN, respectively. We follow these methods to adopt GNN as the building block of our model, and propose a weighted relational graph neural network to overcome the aforementioned limitations.

\section{Problem Formulation}\label{sec:formulation}

In this section, we formulate the problem of POI relationship inference, and
first we introduce some notations to be used. 
Let $\mathcal{P}=\{p_{1},p_{2},...,p_{N}\}$ denote a set of POIs and $\mathcal{C}=\{c_{1},c_{2},...,c_{M}\}$ denote a set of POI categories. 
Each POI $p \in \mathcal{P}$ is associated with a location $l_p$ (longitude and latitude pair), an attribute $x_{p} \in \mathcal{X}$ and a category $c_p \in \mathcal{C}$. POI location and category information can be used to define spatial neighbors and category taxonomy.

\begin{definition}{(\textbf{Spatial Neighbors}).}
Given a distance threshold $d$, the spatial neighbors of a POI $p$ is defined to be $S_{p} := \{p' \in \mathcal{P} \mid \mathrm{dist}(p, p') < d\}$.
\end{definition}

\begin{figure}[tbp]
    \centering
    \includegraphics[width=0.65\linewidth]{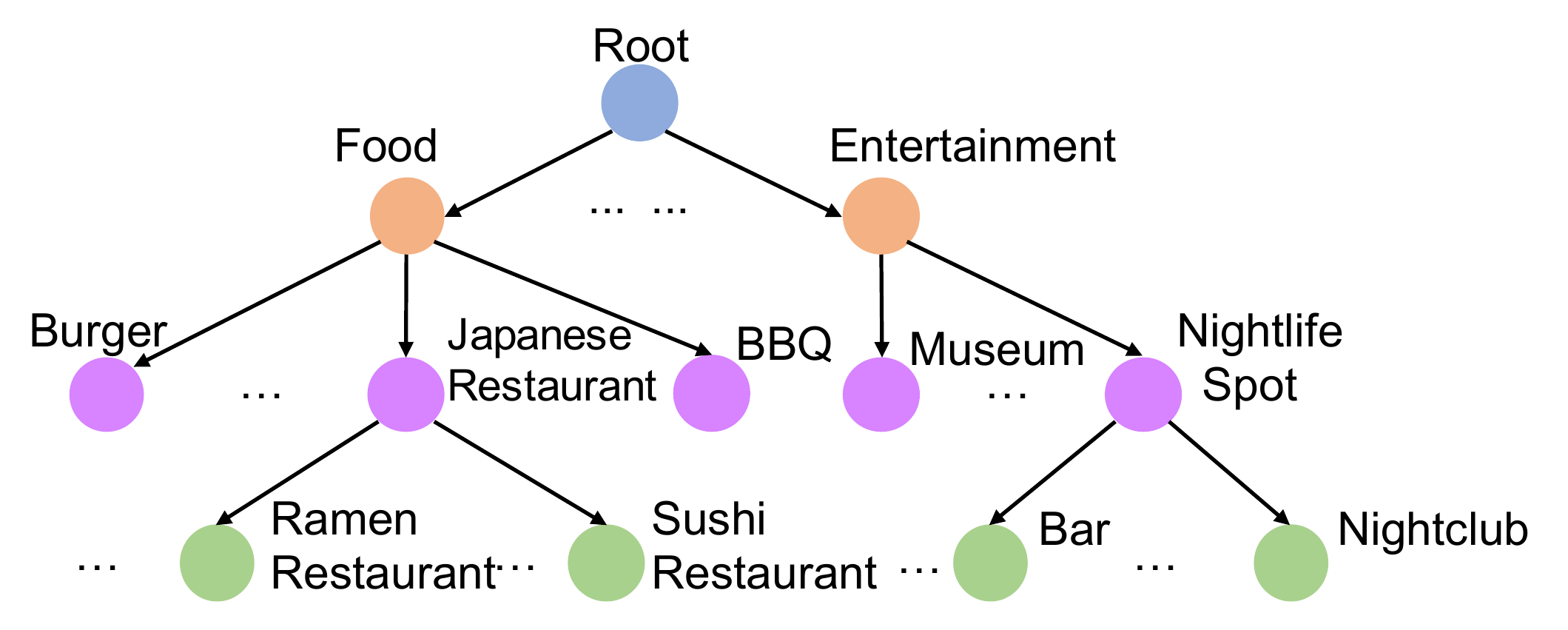}
    \caption{A snapshot of POI category taxonomy}
    \label{fig:taxonomy}
    \vspace{-3mm}
\end{figure}

\begin{definition}{(\textbf{Category Taxonomy}).}
The category taxonomy $T=(\mathcal{C}\cup\mathcal{H},\mathcal{E}_{T})$  is a hierarchical tree, where category set $C$ and hypernym set $\mathcal{H}$ belong to leaf nodes and non-leaf nodes respectively, and each directed edge $e_{t}\in\mathcal{E}_{T}$ represents a hypernymy relation between a category (e.g., food) and its sub-category (e.g., burger). A snapshot of category taxonomy is shown in Figure~\ref{fig:taxonomy}.

\end{definition}

Multiple types of relationships among POIs can be represented as a heterogeneous graph as defined below.

\begin{definition}{(\textbf{Heterogeneous POI Relationship Graph}).}
A heterogeneous POI relationship graph is defined as $G = (\mathcal{P}, \mathcal{E}, \mathcal{R}, \mathcal{X})$.  
Each $p \in \mathcal{P}$ is a POI, and each $e \in \mathcal{E}$ is an edge between a POI pair and associated with a relation type defined by a mapping function $\psi(e):\mathcal{E}\rightarrow\mathcal{R}$; $\mathcal{R}$ is a set of pre-defined semantic relationships between a POI pair.
\end{definition}


Our target is to infer the relationship of a given POI pair that either belongs to a relation type $r\in \mathcal{R}$ or a non-relation type $\phi$. We formally define the problem as below:  

\begin{definition}{(\textbf{POI Relationship Inference Problem}).}
Given a heterogeneous POI relationship graph $G$, a category taxonomy $T$, a distance threshold $d$,
and candidate relation types $\mathcal{R}^{*}=\mathcal{R}\cup\{\phi\}$, we aim to learn a predictive function $f(\mathcal{P}\times\mathcal{P}|G,T,d) \rightarrow \mathcal{R}^{*}$ 
that maps a POI pair $(p_{i},p_{j})$ to a certain relation type.
\end{definition}

In this paper, we focus on two scenarios: (1) $\emph{competitive}$ and $\emph{complementary}$ relationships, as they are of most interests to the business~\cite{WSDM18_itemrelation,KDD15_itemrelation, CIKM20_DecGCN,WSDM19_itemrelation}. POI pairs with competitive relationships refer to those that are interchangeable and provide similar services, while POI pairs with complementary relationships refer to those that tend to be both visited by users; (2) finer-grained multiple relationships that can potentially exist in more complex business situations, such as large-scale spatial knowledge graph completion in the company.

\begin{figure}[tbp]
    \centering
    \includegraphics[width=0.7\linewidth]{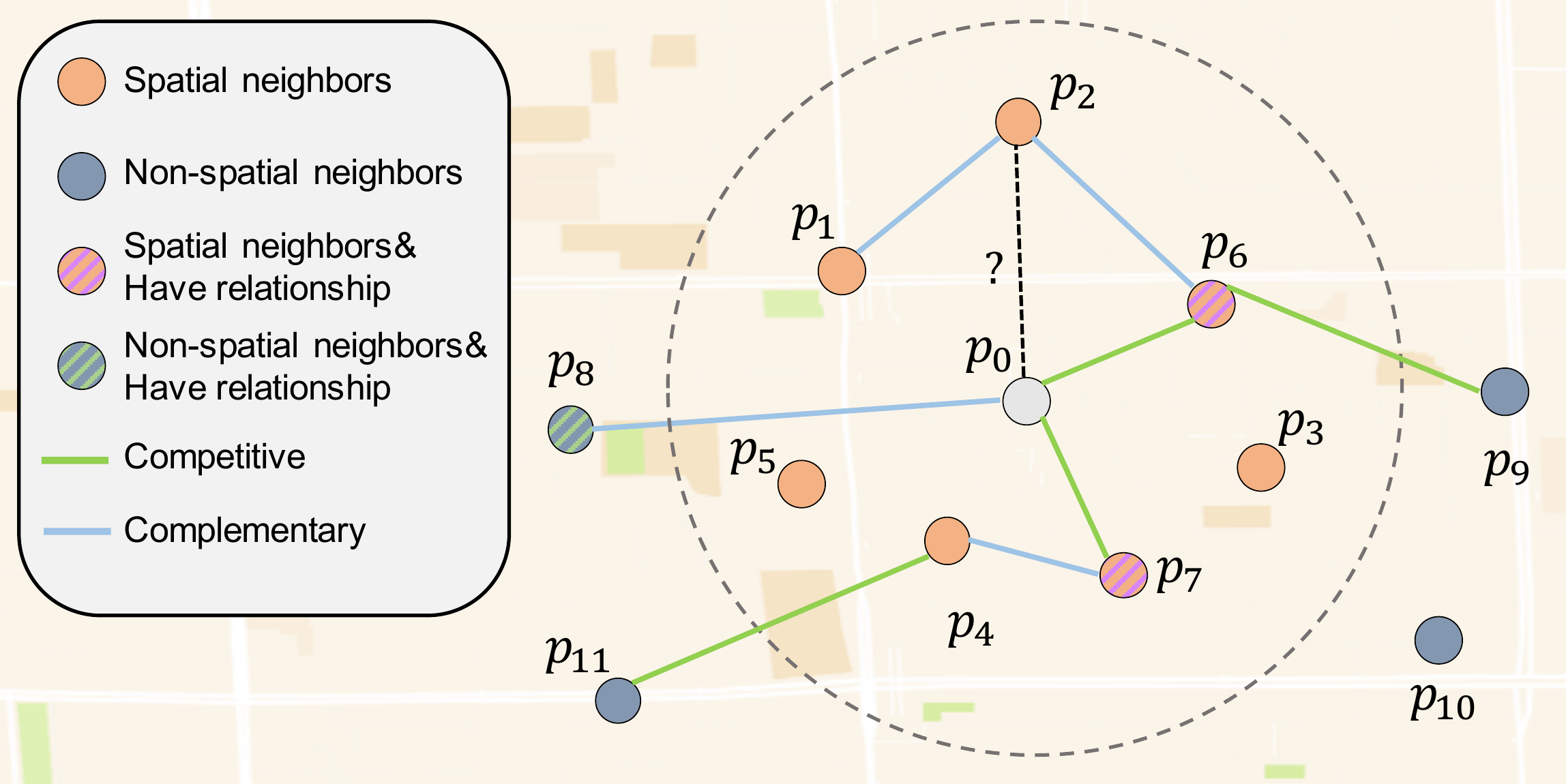}
    \caption{POI relationship inference example}
    \label{fig:example}
    \vspace{-5mm}
\end{figure}

\paratitle{Example.}
As shown in Figure~\ref{fig:example}, $p_{0}$ is the target POI, $p_{1}$-$p_{7}$ which are within the circle are spatial neighbors of $p_0$, $p_{6}$-$p_{8}$ have competitive/complementary relationships with $p_0$ and $p_{8}$ is not the spatial neighbor of $p_{0}$, and $p_{9}$-$p_{11}$ neither are spatial neighbors nor have relationships with $p_{0}$. The inference problem here is to infer the relationship between $p_{0}$ and $p_{2}$. From the example we can observe that spatial neighbors might not have relationships with the target POI (e.g., $p_{3}$,$p_5$), while POIs that have relationships might not be spatial neighbors of the target POI (e.g., $p_8$). In this case, it is important to leverage both of POIs with relationships and spatial neighbors to serve as different aspects of knowledge of a target POI.

\section{Methodology}\label{sec:framework}

\begin{figure*}[htbp]
    \centering
    \includegraphics[width=0.88\linewidth]{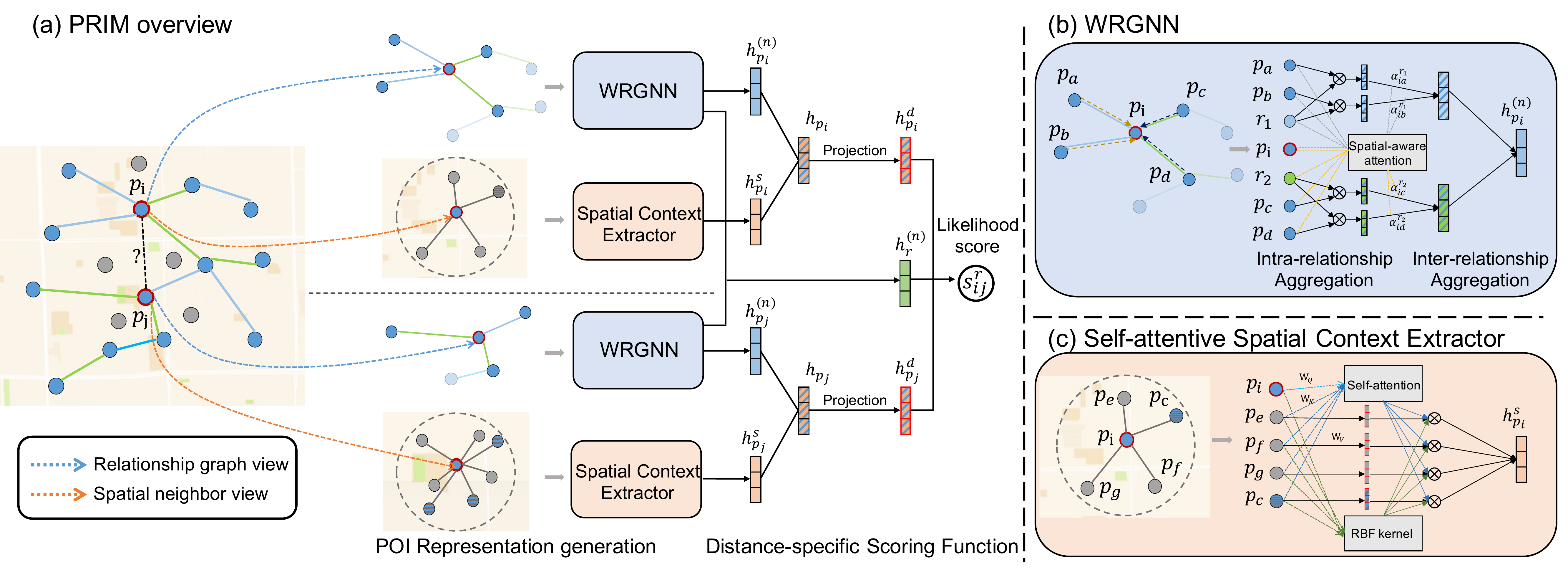}
    \caption{Illustration of the proposed \textsf{PRIM}. (a) the overview architecture of \textsf{PRIM}; (b) the details of WRGNN (we omit the taxonomy integration and relationship representations for clarity); (c) the self-attentive spatial context extractor.}
    \label{fig:framework}
    \vspace{-3mm}
\end{figure*}

We present the four technical components of the proposed \textsf{PRIM}: weighted relational graph neural networks, taxonomy integration module, self-attentive spatial context extractor, and distance-specific scoring objective. Finally, we introduce the training and inference of the model. Figure~\ref{fig:framework} illustrates the proposed framework. 

\subsection{Model Overview}\label{subsec:overview}
Our proposed \textsf{PRIM} consists of four components: weighted relational graph neural network, taxonomy integration module, self-attentive spatial context extractor, and distance-specific scoring function. Figure~\ref{fig:framework} illustrates the proposed framework. Before presenting the details of these components, we discuss the intuitions acquired from the \emph{Beijing} dataset (details in Section~\ref{subsec:dataset}) that guide our module design as well as our technical contributions w.r.t. the design of each component.

First, as discussed in Section~\ref{sec:intro} and ~\ref{sec:relatedwork}, the state-of-the-art GNN based relationship inference methods~\cite{CIKM20_DecGCN,KDD20_POI} and other GNN models~\cite{WWW19_HAN,WWW20_HGT,ICLR20_CompGCN} have at least one of the limitations of (1) failing to handle graph heterogeneity, (2) not capturing spatial characteristics of POIs, (3) ignoring different importance of neighbor nodes. This motivates us to propose a novel weighted relational graph neural network (WRGNN). Specifically, to handle graph heterogeneity, we propose a two-level aggregation process to perform intra-relationship and inter-relationship interactions. Moreover, we design a spatial-aware attention mechanism in which we take into consideration both POI semantic and spatial characteristics to differentiate the importance of neighbor nodes in the aggregation process (Section~\ref{subsec:WRGNN}).

Then, we propose to integrate external structural knowledge, i.e., category taxonomy, to guide the reasoning of WRGNN. Intuitively, POIs whose categories are close in category taxonomy (e.g., \emph{bar} and \emph{nightclub}) tend to be more semantically similar and show higher degree of competitiveness as compared to distant ones (e.g., \emph{bar} and \emph{sushi restaurant}). We calculate the average path distance (i.e., the number of edges along the shortest path between two nodes) on category taxonomy for the corresponding categories of the POI pairs with relationships in our dataset, and we observe that the average path distance is 1.72 for competitive relationships and 3.53 for complementary relationships. Based on this result, we design a simple and effective approach to leverage the 
semantic information in the category taxonomy (Section ~\ref{subsec:taxonomy}).

Next, different from all the previous methods that only focus on modeling the relationship graph, we extract another type of POI representations from spatial neighbors, which provide extra context knowledge about target POIs. This is based on the fact supported by our dataset that we observe some cases where two POI pairs with the same category set or even the same brand set show different relationships when they are located at regions with different spatial context. For example, a KFC and a McDonald, when located in a shopping center, would show less competitiveness than when they are located in a residential area due to larger flow of people. Motivated by this, we propose a self-attentive spatial context extractor to obtain the functionality of the locations where the POIs are distributed, and thus it can help enrich POI representations to encode more information of different aspects  (Section~\ref{subsec:spatial_context}). 

Finally, we argue that the relationship of a POI pair should vary against their spatial distance. For example, given a target restaurant and two identical source restaurants that are $1km$ and $4km$ away to the target restaurant respectively, a distance-agnostic model would yield the same relationship prediction for the two source restaurants. In reality, however, the distant restaurant is less likely to show competitive relationship since competitiveness usually decays with the distance increase. This can be verified in our dataset that 50.1\% of competitive POI pairs are within the distance of $2km$, however the ratio for complementary POI pairs is 21.2\%. Therefore, we propose a distance-specific scoring function to project POI representations to different latent spaces induced by the pairwise distance, thus enhancing the model capacity to capture distance effect of relationships (Section~\ref{subsec:score_func}).

\vspace{-2.5ex}
\subsection{Weighted Relational Graph Neural Network}\label{subsec:WRGNN}

Due to the lack of capacity to model the heterogeneous POI relationship graph in previous methods as discussed in Section~\ref{subsec:overview}, we propose a weighted relational graph neural network (WRGNN) to model graph heterogeneity while considering spatial characteristics in measuring the different importance of neighbors in the aggregation process.

Given a POI $p_{i}$, a two-level aggregation process is adopted. Specifically, neighbors connected by the same relation type are first aggregated (intra-relationship aggregation), and then results from different types of relationships are further aggregated (inter-relationship aggregation) to generate the final representation for $p_{i}$:
\begin{equation}\label{WRGNN1}
    \mathbf{h}_{p_{i}}^{(l+1)} = \sigma \left(\sum_{r\in\mathcal{R}}\sum_{p_{j}\in\mathcal{N}_{p_{i}}^{r}}\alpha_{ij}^{r}\mathbf{W}^{(l)}\gamma(\mathbf{h}^{(l)}_{p_{j}},\mathbf{h}_{r}^{(l)})\right)
\end{equation}
Here, $\mathbf{h}_{p_{i}}^{(l)}$, $\mathbf{h}_{r}^{(l)}$ are the representations of $p_{i}$ and relationship of type $r$ in the $l$-th layer, respectively, $\mathcal{N}_{p_{i}}^{r}$ denotes a set of graph neighbors of $p_{i}$ with relationship of type $r$, $\alpha_{ij}^{r}$ is normalized importance score of $p_{j}$ to $p_{i}$, $\mathbf{W}^{(l)}$ is the weight matrix in the $l$-th layer, $\gamma(.,.)$ is a relation-specific operation, and $\sigma(.)$ is an activation function.

The relation-specific operator $\gamma(.,.)$ takes a POI-relationship pair as input and allows the interaction between a POI and a relationship in the aggregation process. It has several advantages: (1) we jointly learn POI and relationship representations, which can be seamlessly applied to the proposed scoring function (Section~\ref{subsec:score_func}); (2) relation semantics are considered so that a neighbor POI would have different influence on the target POI for each relation type; (3) $\gamma(.,.)$ is quite flexible and can be chosen from various options, such as multiplication~\cite{ICLR_mult}, circular-correlation~\cite{AAAI16_corr}, and complex neural network based operations~\cite{NIPS_NTN}. Here we choose element-wise multiplication $\gamma(\mathbf{h}^{(l)}_{p_{j}},\mathbf{h}_{r}^{(l)})=\mathbf{h}^{(l)}_{p_{j}}\odot \mathbf{h}_{r}^{(l)}$ due to its efficiency and comparable results to other options. 


Furthermore, the relationship representation of $r\in\mathcal{R}$ is updated as follows:
\begin{equation}
    \mathbf{h}_{r}^{(l+1)} = \mathbf{W}_{r}^{(l)}\mathbf{h}_{r}^{(l)}
\end{equation}
where $\mathbf{W}_{r}^{(l)}$ is a weight matrix in the $l$-th layer for all the relationships. As a result of these two updating functions, POI and relationship representations are refined alternatively by stacking multiple such layers. By doing this, these two types of representations are highly coupled, and thus can complement each other to learn rich semantic characteristics from the heterogeneous POI relationship graph. Such a design makes WRGNN more capable of handling graph heterogeneity than those methods that model each relation type separately through decomposition~\cite{KDD20_POI,CIKM20_DecGCN}.

In previous methods~\cite{RGCN, ICLR20_CompGCN, KDD20_POI}, neighbors in each relationship are treated equally and $\alpha_{ij}^{r}$ is set to be $\frac{1}{|N_{p_{i}}^{r}|}$. Some GNN models have shown the capability of attention mechanism on deriving the importance of neighbors by automatically reasoning their features~\cite{WWW20_MAGNN,WWW20_HGT}. Inspired by them, we propose a spatial-aware attention mechanism to learn the importance score $\alpha_{ij}^{r}$ of graph neighbors by combining the POI semantic and spatial characteristics together as follows:

\begin{align}
    e_{i j}^{r}&=\sigma\left(\mathbf{a}_{r}^{\top} \cdot\left[\mathbf{W}_{a} \mathbf{h}_{p_{i}}^{(l)}\left\|\mathbf{W}_{a} \mathbf{h}_{p_{j}}^{(l)}\right\| \mathbf{W}_{d} \mathbf{d}_{i j}\right]\right), \label{WRGNN_att}\\
   \alpha_{i j}^{r}&=\operatorname{softmax}\left(e_{i j}^{r}\right)=\frac{\exp \left(e_{i j}^{r}\right)}{\sum_{k \in \mathcal{N}_{p_{i}}^{r}} \exp \left(e_{i k}^{r}\right)},
\end{align}
 where $\mathbf{a}_{r}$ is the attention vector shared for the neighbors with respect to the relationship $r$, $\mathbf{W}_{a}$ is the transformation matrix applied to POI representations,  $\mathbf{W}_{d}$ is the transformation matrix applied to spatial distance $\mathbf{d}_{ij}$ between $p_{i}$ and $p_{j}$, $\|$ denotes the vector concatenation operation, $\sigma()$ is an activation function, and $e_{i j}^{r}$ is normalized by a softmax function to get $\alpha_{ij}^{r}$.

As suggested in~\cite{GAT, NIPS_attn}, the above attention mechanism can be extended to multiple heads,  which helps stabilize the learning process. Specifically, $K$ independent attention mechanisms are executed and then concatenated to form the output POI representations. We re-write Equation~\ref{WRGNN1} as follows:

\begin{equation}\label{WRGNN_agg}
    \mathbf{h}_{p_{i}}^{(l+1)} = \overset{K}{\underset{k=1}{\big| \big|}}\sigma\left(\sum_{r\in\mathcal{R}}\sum_{p_{j}\in\mathcal{N}_{p_{i}}^{r}}\alpha_{ij,k}^{r}\mathbf{W}_{k}^{(l)}\gamma(\mathbf{h}^{(l)}_{p_{j}},\mathbf{h}_{r}^{(l)})\right)
\end{equation}
where $\alpha_{ij,k}^{r}$ is the normalized importance score of $p_{j}$ to $p_{i}$ with respect to relationship $r$ in the $k$-th attention head, $\|$ denotes the concatenated representation from $K$ attention heads.

\subsection{Taxonomy Integration}\label{subsec:taxonomy}

We propose to integrate category taxonomy as auxiliary structural knowledge to guide the learning of WRGNN. As shown in Figure~\ref{fig:taxonomy}, the categories are organized into different levels to present category concepts at different degrees of granularity, from general to specific. Category taxonomy provides explicit concept similarity for different categories that can be utilized to enrich the learned POI representations from WRGNN.


As discussed in Section~\ref{subsec:overview}, we adopt a simple yet effective approach to encode category taxonomy. Here we use a POI $p_{i}$ as an example to illustrate.
Specifically, each node $t \in T$ in the category taxonomy tree is embedded into a vector $\mathbf{e}_t$. Then given POI $p_{i}$, we backtrack to retrieve all the nodes from its corresponding leaf node to the root node in $T$, denoted by $Q_{p_{i}}$. For example, the category path for \emph{bar} is [\emph{root}, \emph{entertainment}, \emph{nightlife spot}, \emph{bar}]. Then we derive the category representation $\mathbf{q}_{p_{i}}$ for $p_{i}$ as follows: $\mathbf{q}_{p_{i}} = \sum_{t \in Q_{p_{i}}} \mathbf{e}_{t}$

In this case, close categories would share more common elements in the category path, thus resulting in more similar representations. After that, we concatenate the original representations generated from WRGNN and the category representations defined as: $\mathbf{h}_{p_{i}}^{*(l)} =[\mathbf{h}_{p_{i}}^{(l)} \| \mathbf{q}_{p_{i}}]$. 
Afterwards, we integrate the taxonomy by simply replacing the original POI representation $\mathbf{h}_{p_{i}}^{(l)}$ with $\mathbf{h}_{p_{i}}^{*(l)}$ in Equation~\ref{WRGNN_att} \&~\ref{WRGNN_agg} and adjusting the dimension of parameters correspondingly. Note that we do not apply more complicated techniques, such as Tree-LSTM~\cite{Tree-LSTM} and hyperbolic embedding~\cite{hyperbolic}, to model category taxonomy because they are less efficient and found to show no improvement over the proposed solution in our preliminary experiments.

\subsection{Self-attentive Spatial Context Extractor}\label{subsec:spatial_context}
The modules introduced so far focus on modeling a target POI based on its neighbors in the heterogeneous POI relationship graph. Apart from graph neighbors whose number is usually limited, a POI usually has a large number of spatial neighbors that are overlooked in WRGNN. They can be leveraged to provide extra spatial context that is useful in inferring the relationship between POIs as discussed in Section~\ref{subsec:overview}.

We propose to generate spatial context as another view of POI representations. Specifically, we 
apply the self-attention~\cite{NIPS_attn} technique where a target POI is treated as a query and its spatial neighbors are treated as a set of key-value pairs. 
Given the learned POI representations from WRGNN, we calculate the spatial context $\mathbf{h}_{p_{i}}^{s}$ of POI $p_{i}$ as follows:
\begin{equation}
\begin{aligned}
    \mathbf{h}_{p_{i}}^{s} = \sum_{p_{j}\in \mathcal{S}_{p_{i}}}&\beta_{ij} (\mathbf{W}_{V}\mathbf{h}_{p_{j}}^{(L)}) \\ 
     \beta_{ij} = \operatorname{softmax} ( e_{i j}&)= \frac{\exp \left(e_{ij}\right)}{\sum_{p_{m}\in \mathcal{S}_{p_{i}}} \exp \left(e_{i m}\right)}
\end{aligned}
\end{equation}
Here, $\mathbf{h}_{p_{j}}^{(L)}$ is the output after $L$ layers of WRGNN for $p_{j}$, $\mathcal{S}_{p_{i}}$ is the spatial neighbors of $p_{i}$, $\beta_{ij}$ is the normalized weight derived from $e_{ij}$ via softmax function which indicates the attention score between $p_{i}$ and $p_{j}$, and $\mathbf{W}_{V}$ is the projection matrix for values. 

To derive the attention scores, the target POI is regarded as a query to attend over all the spatial neighbors which are regarded as keys, and the attention score for one spatial neighbor $p_{j}$ can be expressed as follows:

\begin{equation}\label{self_att}
    e_{ij}^{\prime} = \frac{\left(\mathbf{W}_{Q} \mathbf{h}_{p_{i}}^{(L)}\right)^{\top} \cdot\left(\mathbf{W}_{K} \mathbf{h}_{p_{j}}^{(L)}\right)}{\sqrt{d_{p}}}
\end{equation}
where $\mathbf{W}_{Q}$ and $\mathbf{W}_{K}$ are the projection matrices for query and keys respectively, and $d_{p}$ is the dimension of the POI representations. However, it does not explicitly take the geographical influence into consideration. According to the First Law of Geography~\cite{GeoLaw}, everything is related to everything else, but near things are more related than distant things. It indicates that more emphasis should be put on nearby spatial neighbors than on farther ones to extract spatial context. Therefore, we use the radial basis function (RBF) kernel to assign weights for spatial neighbors as follows:
\begin{equation}\label{geo_att}
D\left(l_{p_{i}}, l_{p_{j}}\right)=\exp \left( -\theta\left\|l_{p_{i}}-l_{p_{j}}\right\|^{2}\right)    
\end{equation}
where $l_{p_{i}}$ and $l_{p_{j}}$ are location coordinates of $p_{i}$ and $p_{j}$ respectively and $\theta$ is a scaling parameter. After that, we combine Equation~\ref{self_att} and~\ref{geo_att} to derive the final attention scores which consider both semantic information and geographical influence for POIs by multiplication:
\begin{equation}
    e_{ij} = e_{ij}^{\prime}\cdot D\left(l_{p_{i}}, l_{p_{j}}\right)  
\end{equation}

Thus far, we have obtained the representations of POI from two views, namely the \emph{heterogeneous POI relationship graph view} and the \emph{spatial context view}. For each $p_{i}\in\mathcal{P}$, we fuse these two views to get the final POI representation by:

\begin{equation}
     {\mathbf{h}}_{p_{i}} = \mathbf{h}_{p_{i}}^{(L)}+\mathbf{h}_{p_{i}}^{s} 
\end{equation}

\subsection{Distance-specific Scoring Function}\label{subsec:score_func}
We have obtained the relationship representations and the final POI representations. Now we propose a distance-specific scoring function based on the intuition discussed in Section~\ref{subsec:overview}.

First, we project the learned POI representations to a distance-specific hyperplane according to the involved POI pairs. 
Specifically, we split the distance into non-overlapping bins (e.g., $0$-$1km$, $1$-$2km$, etc) and assign each bin with a hyperplane specified by a unit normal vector $\mathbf{w}_{b}$. Then, given the POI pair $(p_{i}, p_{j}) \in \mathcal{P} \times \mathcal{P}$ and their final representations, we project them into the hyperplane specified by distance as follows:

\begin{equation}\label{hyper}
   \mathbf{h}_{p_{i}}^{d}=\mathbf{h}_{p_{i}}- \mathbf{w}_{g(d_{ij})}^{\top}\mathbf{w}_{g(d_{ij})} \mathbf{h}_{p_{i}}
\end{equation}
where $\mathbf{w}_{g(d_{ij})}$ is the unit normal vector of the bin to which the POI pair belongs, $g(d_{ij})$ is a look-up function to map the distance $d_{ij}$ between $p_{i}$ and $p_{j}$ to its bin, and ${\mathbf{h}}_{p_{i}}^{d}$ is the projected representation for $p_{i}$.

Next, we compute the likelihood of different relationships based on the projected POI representations and relationship representations. Since the symmetric property holds for the relationship of a POI pair, we adopt the scoring function in~\cite{ICLR_mult} to capture this property. For a POI pair $(p_{i},p_{j})$ and a relation type $r$, it is defined as follows:
\begin{equation}
    s_{i j}^{r}=\mathbf{h}_{p_{i}}^{d \top} \operatorname{diag}\left(\mathbf{h}_{r}^{(L)}\right) \mathbf{h}_{p_{j}}^{d}
\end{equation}
where $\mathbf{h}_{r}^{(L)}$ is the representation for relation type $r\in\mathcal{R}^{*}$ after $L$ layers of WRGNN, and $s_{ij}^{r}$ is the likelihood score of $p_{i}$ and $p_{j}$ having relationship $r$.

\subsection{Training and Inference}
We train the model via the cross-entropy loss with negative sampling~\cite{NS}. For each observed positive triplet $(p_{i}, r, p_{j})$ that denotes a relationship $r$ exists between POI pair $(p_{i},p_{j})$, we sample $\omega$ negative pairs by replacing a POI with a randomly sampled one. Then the loss function can be expressed as follows:

\begin{equation}
     \mathcal{L} = \sum_{(p_{i},r,p_{j})\in\mathcal{D}}y_{ij}^{r}\log\sigma(s_{ij}^{r}) + (1-y_{ij}^{r})\log\sigma(1-s_{ij}^{r})
\end{equation}
where $\mathcal{D}$ is the total set of positive and negative triples, $\sigma$ is the sigmoid function, and $y_{ij}^{r}$ is the label indicator which is set to 1 for positive triples and 0 for negative ones.

During inference, given a pair of $(p_{i},p_{j})$, we calculate the likelihood score w.r.t. each relation type. Then we rank the scores and select the relationship with the highest score as the prediction result: $\hat{r}_{ij} = \arg\max_{r\in\mathcal{R}^{*}}s_{ij}^{r}$. It is worth noting that our model has the ability of inductive reasoning and can be applied in unseen cases. This feature is desirable in practical use since the model does not require being frequently updated for newly arrived POIs and is scalable to large datasets.


\paratitle{Time complexity.}
We assume that the POI relationship graph consists of $m$ edges and $n$ POIs, the dimension of POI representation is $d$, and the average spatial neighbors for each POI is $\Tilde{S}$. The time complexity of WRGNN (Sec 4.2-4.3), spatial context extractor (Sec 4.4) and distance-specific scoring function (Sec 4.5) for all the POIs are $O(Lnd^{2}+Lmd)$, $O(nd^{2}+n\Tilde{S}d)$ and $O(nd)$, respectively. Since $m\gg n$ and other parameters can be considered as constants, $O(Lmd)$ is the dominating complexity term that grows linearly with the edge number.

\section{Experiments}\label{sec:exp}

In this section, we study the performance of the proposed model on two real-world datasets.

\subsection{Experimental Settings}\label{subsec:dataset}
\subsubsection{Datasets}
We evaluate all methods using two real-world city-wide datasets, $\emph{Beijing}$ (BJ) and $\emph{Shanghai}$ (SH), from Meituan\footnote[1]{https://www.meituan.com} which is one of the largest location-based service providers in China. We construct relationships between POIs from user logs extracted from these two datasets. Following the previous work~\cite{KDD15_itemrelation, WSDM18_itemrelation, CIKM20_DecGCN}, we generate the ground truth of $\emph{competitive}$ relationship for a POI pair if the two POIs in the pair are "viewed/clicked together" by users within a query session, and we generate the ground truth of $\emph{complementary}$ relationships for a POI pair if the two POIs in the pair are  "also viewed/clicked" by same users across different query sessions. Note that user logs 
are often unavailable or insufficient, and thus the way of generating ground-truth cannot be generalized to cities where they are unavailable, or are insufficient to cover all the POIs. 
Our proposed methods do not use user logs, i.e., the information of generating the ground-truth relationships for evaluation, and are applicable when the user logs are unavailable, or can be used to enrich the POI relationships when the initial data sources are incomplete. Details of the extracted graphs are shown in Table~\ref{tab:dataset}.

\begin{table}[htp]
\centering
\caption{Statistics of the datasets}
\vspace{-2mm}
\resizebox{0.9\linewidth}{!}{
\begin{tabular}{c|c|c|c|c}
\hline
\multirow{2}{*}{Dataset} & \multicolumn{2}{c|}{Taxonomy}              & \multirow{2}{*}{\#POIs} & \multirow{2}{*}{\#Relational Edges} \\ \cline{2-3}
                         & \#Non-leaf nodes    & \#Categories         &                         &                                     \\ \hline
\emph{Beijing}                  & \multirow{2}{*}{95} & \multirow{2}{*}{805} & 13,334                  & 122,462                             \\ \cline{1-1} \cline{4-5} 
\emph{Shanghai}                 &                     &                      & 10,090                  & 112,848                             \\ \hline
\end{tabular}}
\label{tab:dataset}
\vspace{-3mm}
\end{table}

\subsubsection{Baselines and Evaluation Metrics}
We compare our model with five types of baseline methods, including rule-based methods ($\textbf{CAT}$, $\textbf{CAT-D}$), random walk based graph embedding methods ($\textbf{Deepwalk}$, $\textbf{node2vec}$), vanilla GNN ($\textbf{GCN}$, $\textbf{GAT}$), heterogeneous GNN ($\textbf{HAN}$, $\textbf{HGT}$, $\textbf{R-GCN}$, $\textbf{CompGCN}$), and the state-of-the-art relationship inference methods ($\textbf{DecGCN}$, $\textbf{DeepR}$). 
\begin{itemize}
    \item $\textbf{CAT}$ and $\textbf{CAT-D}$. CAT determines the relation type by a threshold of POI category distance on the category taxonomy, and CAT-D considers both POI geographical distance and POI category distance thresholds. We search the thresholds that achieve the best results on two datasets.
    \item $\textbf{Deepwalk}$~\cite{Deepwalk} and $\textbf{node2vec}$~\cite{node2vec} learn node representations from sequences generated by random walks on a graph. node2vec adopts biased random walks, which is more generalized compared to the unbiased ones adopted in Deepwalk.
    \item $\textbf{GCN}$~\cite{GCN} and $\textbf{GAT}$~\cite{GAT} are two vanilla GNN models which produce representations by aggregating from neighbor nodes. GCN treats neighbor nodes equally while GAT considers different weights for them via attention mechanism. These two methods do not model different relation types between POIs.
    \item $\textbf{HAN}$~\cite{WWW19_HAN} is a meta-path based heterogeneous GNN model, which performs node-level and semantic-level attention to model multiple relation types.  
    \item $\textbf{HGT}$~\cite{WWW20_HGT} handles graph heterogeneity with relation-specific mutual attention mechanism which models neighbor aggregation in different latent spaces for each relation type.
    \item $\textbf{R-GCN}$~\cite{RGCN} is an extension of GCN to better model relational graphs by assigning different weight matrices for different relation types in neighbor aggregation.
    \item $\textbf{CompGCN}$~\cite{ICLR20_CompGCN} jointly learns the embeddings of nodes and relations for heterogeneous graph and leverages multiple composition functions to update the node representations.
    \item $\textbf{DecGCN}$~\cite{CIKM20_DecGCN} is a state-of-the-art product relationship inference method that decomposes a heterogeneous graph into sub-graphs, one for each relation type. GNN is applied in each sub-graph followed by a co-attention mechanism to extract supplementary information from other sub-graphs. 
    \item $\textbf{DeepR}$~\cite{KDD20_POI} is a state-of-the-art method that aims to infer the competitiveness between POIs. It considers spatial features by splitting the neighbors of a POI into different sectors based on the coordinates and performs aggregation from each sector. To make it applicable to multiple relationships, we extract a sub-graph for each relation type, and apply the method to each sub-graph.  
\end{itemize}

To evaluate the performance of different methods, we adopt the widely-used \emph{Micro-F1} and \emph{Macro-F1} in previous studies~\cite{WSDM18_itemrelation,KDD20_POI} to jointly consider precision and recall. A higher score indicates a better performance.

\subsubsection{Parameter Settings}
In our experiments, we randomly sample 10\% of the edges as validation data, 20\% of the edges as test data, and different fractions of the remaining edges are used as training data. Apart from the relational edges, 16,000 POI pairs which do not have relationships are randomly sampled for the non-relation type for testing. In our model, we set the threshold $d$ for determining spatial neighbors to be $1.15km$, the scaling factor $\theta$ in the radial basis function (RBF) kernel to be 2, the number of heads to be 4, the category embedding size to be 128, and the number of negative samples $\omega$ to be 5. For random walk based models, we set window size to 5, walk length to 30, and walks per node to 20. For all the GNN based methods, we set the number of layers to be 3. To achieve fair comparison, POI embedding size is set to be 128 for all the methods. We use Adam optimizer with learning rate of 0.001 and batch size of 512 to train our model.

\begin{table*}[tbp]
\centering
\caption{Results on the two datasets in terms of Macro-F1 and Micro-F1 (with best in bold and second-best underlined)}
\vspace{-3mm}
\resizebox{0.9\linewidth}{!}{
\begin{tabular}{|c|c|c||c|c|c|c|c|c|c|c|c|c|c|c|c|}
\hline
Dataset                    & Metric                    & Train\% & CAT   & CAT-D & Deepwalk & node2vec & GCN   & GAT   & HAN   & HGT   & R-GCN & CompGCN & DecGCN & DeepR & PRIM  \\ \hline
\multirow{8}{*}{BJ}  & \multirow{4}{*}{Macro-F1} & 40\%    & 0.464 & 0.519 & 0.638    & 0.640    & 0.707 & 0.724 & 0.782 & 0.779 & 0.789 & \underline{0.794}   & 0.757  & 0.783 & \textbf{0.845} \\ 
                           &                           & 50\%    & 0.464 & 0.519 & 0.691    & 0.692    & 0.737 & 0.748 & 0.811 & 0.814 & 0.814 & \underline{0.832}   & 0.801  & 0.815 & \textbf{0.870} \\ 
                           &                           & 60\%    & 0.464 & 0.519 & 0.731    & 0.734    & 0.755 & 0.776 & 0.839 & 0.842 & 0.820 & \underline{0.860}   & 0.811  & 0.842 & \textbf{0.882} \\  
                           &                           & 70\%    & 0.464 & 0.519 & 0.757    & 0.761    & 0.770 & 0.795 & 0.857 & 0.857 & 0.828 & \underline{0.870}   & 0.823  & 0.861 & \textbf{0.895} \\ \cline{2-16} 
                           & \multirow{4}{*}{Micro-F1} & 40\%    & 0.559 & 0.579 & 0.707    & 0.710    & 0.729 & 0.753 & 0.817 & 0.813 & 0.808 & \underline{0.827}   & 0.805  & 0.820 & \textbf{0.879} \\ 
                           &                           & 50\%    & 0.559 & 0.579 & 0.758    & 0.762    & 0.766 & 0.776 & 0.842 & 0.845 & 0.827 & \underline{0.859}   & 0.823  & 0.847 & \textbf{0.895} \\ 
                           &                           & 60\%    & 0.559 & 0.579 & 0.783    & 0.784    & 0.780 & 0.804 & 0.867 & 0.869 & 0.832 & \underline{0.882}   & 0.826  & 0.871 & \textbf{0.907} \\ 
                           &                           & 70\%    & 0.559 & 0.579 & 0.816    & 0.817    & 0.796 & 0.821 & 0.882 & 0.883 & 0.839 & \underline{0.892}   & 0.843  & 0.887 & \textbf{0.913} \\ \hline
                           
\multirow{8}{*}{SH} & \multirow{4}{*}{Macro-F1} & 40\%    & 0.443 & 0.509 & 0.652    & 0.655    & 0.673 & 0.692 & 0.788 & \underline{0.790} & 0.777 & 0.776   & 0.765  & 0.786 & \textbf{0.822} \\ 
                           &                           & 50\%    & 0.443 & 0.509 & 0.682    & 0.684    & 0.704 & 0.711 & 0.807 & \underline{0.811} & 0.802 & 0.808   & 0.797  & 0.808 & \textbf{0.844} \\ 
                           &                           & 60\%    & 0.443 & 0.509 & 0.698    & 0.702    & 0.713 & 0.729 & 0.834 & 0.838 & 0.828 & 0.837   & 0.817  & \underline{0.839} & \textbf{0.861} \\  
                           &                           & 70\%    & 0.443 & 0.509 & 0.707    & 0.711    & 0.731 & 0.735 & 0.853 & \underline{0.858} & 0.839 & 0.857   & 0.832  & 0.852 & \textbf{0.875} \\ \cline{2-16} 
                           & \multirow{4}{*}{Micro-F1} & 40\%    & 0.551 & 0.573 & 0.724    & 0.727    & 0.744 & 0.769 & 0.843 & 0.841 & 0.822 & 0.836   & 0.824  & \underline{0.845} & \textbf{0.886} \\  
                           &                           & 50\%    & 0.551 & 0.573 & 0.766    & 0.768    & 0.775 & 0.788 & 0.861 & \underline{0.864} & 0.837 & 0.860   & 0.846  & 0.863 & \textbf{0.896} \\ 
                           &                           & 60\%    & 0.551 & 0.573 & 0.783    & 0.786    & 0.793 & 0.806 & 0.886 & 0.888 & 0.864 & \underline{0.891}   & 0.856  & 0.890 & \textbf{0.909} \\ 
                           &                           & 70\%    & 0.551 & 0.573 & 0.796    & 0.798    & 0.804 & 0.811 & 0.897 & 0.901 & 0.872 & \underline{0.903}   & 0.869  & 0.898 & \textbf{0.920} \\ \hline
                    
\end{tabular}}
\label{tab:result}
\vspace{-2mm}
\end{table*}

\begin{table*}[tbp]
\centering
\caption{Results on multiple relationships in terms of Macro-F1 and Micro-F1 (with best in bold and second-best underlined)}
\vspace{-3mm}
\resizebox{0.75\linewidth}{!}{
\begin{tabular}{|c|c|c||c|c|c|c|c|c|c|c|c|c|}
\hline
Dataset             & Metric                    & Train\% & Deepwalk & node2vec & GCN   & GAT   & HAN   & HGT   & R-GCN & CompGCN & DeepR & PRIM  \\ \hline
\multirow{8}{*}{BJ} & \multirow{4}{*}{Macro-F1} & 40\%    & 0.475    & 0.477    & 0.579 & 0.577 & 0.605 & 0.620 & 0.591 & 0.603   & \underline{0.626} & \textbf{0.664} \\ 
                    &                           & 50\%    & 0.518    & 0.522    & 0.611 & 0.603 & 0.641 & 0.654 & 0.632 & 0.647   & \underline{0.656} & \textbf{0.678} \\ 
                    &                           & 60\%    & 0.543    & 0.544    & 0.625 & 0.621 & 0.660 & 0.675 & 0.657 & \underline{0.676}   & 0.673 & \textbf{0.694} \\ 
                    &                           & 70\%    & 0.576    & 0.579    & 0.632 & 0.626 & 0.688 & \underline{0.701} & 0.676 & 0.687   & 0.697 & \textbf{0.721} \\ \cline{2-13} 
                    & \multirow{4}{*}{Micro-F1} & 40\%    & 0.638    & 0.637    & 0.673 & 0.675 & 0.678 & \underline{0.716} & 0.680 & 0.699   & 0.714 & \textbf{0.759} \\ 
                    &                           & 50\%    & 0.667    & 0.669    & 0.696 & 0.691 & 0.717 & \underline{0.744} & 0.709 & 0.716   & 0.742 & \textbf{0.789} \\ 
                    &                           & 60\%    & 0.689    & 0.693    & 0.718 & 0.714 & 0.724 & \underline{0.761} & 0.721 & 0.751   & 0.752 & \textbf{0.799} \\ 
                    &                           & 70\%    & 0.709    & 0.711    & 0.723 & 0.722 & 0.756 & 0.776 & 0.755 & 0.764   & \underline{0.778} & \textbf{0.804} \\ \hline
\multirow{8}{*}{SH} & \multirow{4}{*}{Macro-F1} & 40\%    & 0.472    & 0.471    & 0.474 & 0.473 & 0.524 & \underline{0.547} & 0.532 & 0.531   & 0.543 & \textbf{0.582} \\ 
                    &                           & 50\%    & 0.503    & 0.505    & 0.504 & 0.506 & 0.556 & \underline{0.579} & 0.555 & 0.559   & 0.576 & \textbf{0.604} \\
                    &                           & 60\%    & 0.520    & 0.523    & 0.526 & 0.523 & 0.601 & 0.622 & 0.597 & 0.575   & \underline{0.623} & \textbf{0.642} \\  
                    &                           & 70\%    & 0.533    & 0.537    & 0.542 & 0.529 & 0.613 & 0.631 & 0.614 & 0.611   & \underline{0.634} & \textbf{0.659} \\ \cline{2-13} 
                    & \multirow{4}{*}{Micro-F1} & 40\%    & 0.635    & 0.638    & 0.627 & 0.622 & 0.676 & \underline{0.718} & 0.688 & 0.678   & 0.713 & \textbf{0.753} \\ 
                    &                           & 50\%    & 0.657    & 0.659    & 0.674 & 0.664 & 0.714 & 0.742 & 0.694 & 0.694   & \underline{0.744} & \textbf{0.778} \\ 
                    &                           & 60\%    & 0.677    & 0.681    & 0.679 & 0.669 & 0.744 & \underline{0.761} & 0.742 & 0.746   & 0.756 & \textbf{0.790} \\
                    &                           & 70\%    & 0.702    & 0.704    & 0.694 & 0.695 & 0.752 & \underline{0.778} & 0.768 & 0.771   & 0.774 & \textbf{0.806} \\ \hline
\end{tabular}}
\label{tab:result_multi}
\end{table*}

\vspace{-2ex}
\subsection{Performance Comparison}
The results of different methods are reported in Table~\ref{tab:result}, where we use different fractions of datasets from 40\% to 70\% as training data (denoted by ``Train\%''). We have several observations: (1) Rule-based methods, random walk based methods and vanilla GNN methods perform poorly on both datasets, demonstrating that simple rules and methods which do not consider relational features are not sufficient to provide accurate predictions. (2) Methods that can model the relation heterogeneity achieve relatively good results as compared to other baselines. Moreover, the methods that decompose a graph into sub-graphs for different relation types (DecGCN, DeepR) perform worse than those that model multiple relation types within a graph in a unified way (HGT, CompGCN). (3) HGT and CompGCN outperform other baselines in most cases, which affirms the effectiveness of attention mechanism in aggregation from neighbor nodes and learning relation representations. (4) Our proposed \textsf{PRIM} outperforms all the baselines in all metrics and in all cases. Compared to HGT and CompGCN, PRIM captures unique spatial features of POIs with novel spatial-aware attention mechanism and spatial context extractor, and employs distance-specific scoring function to better utilize relation representations. 

To further validate the effectiveness of handling multiple relation types, based on the number of times appearing in user logs, we further create finer-grained relationships to differentiate the degree of competitiveness/complementarity between POI pairs. Finally, we end up with 6 relationships and the experimental results for GNN  based models are shown in Table~\ref{tab:result_multi}. It can be observed that the proposed $\textsf{PRIM}$ still performs the best, which further demonstrates the superiority of our model.

\subsection{Model Scalability}
Given that the above datasets are relatively small, we study the model scalability on a large POI dataset in Singapore that consists of 251,219 POIs. Due to the lack of ground-truth relationships, for each POI we randomly assign 8 relationships to others. We evaluate the scalability for POIs ranging from 50K to 250K.

\begin{figure}[htbp]
    \centering
    \includegraphics[width=0.7\linewidth]{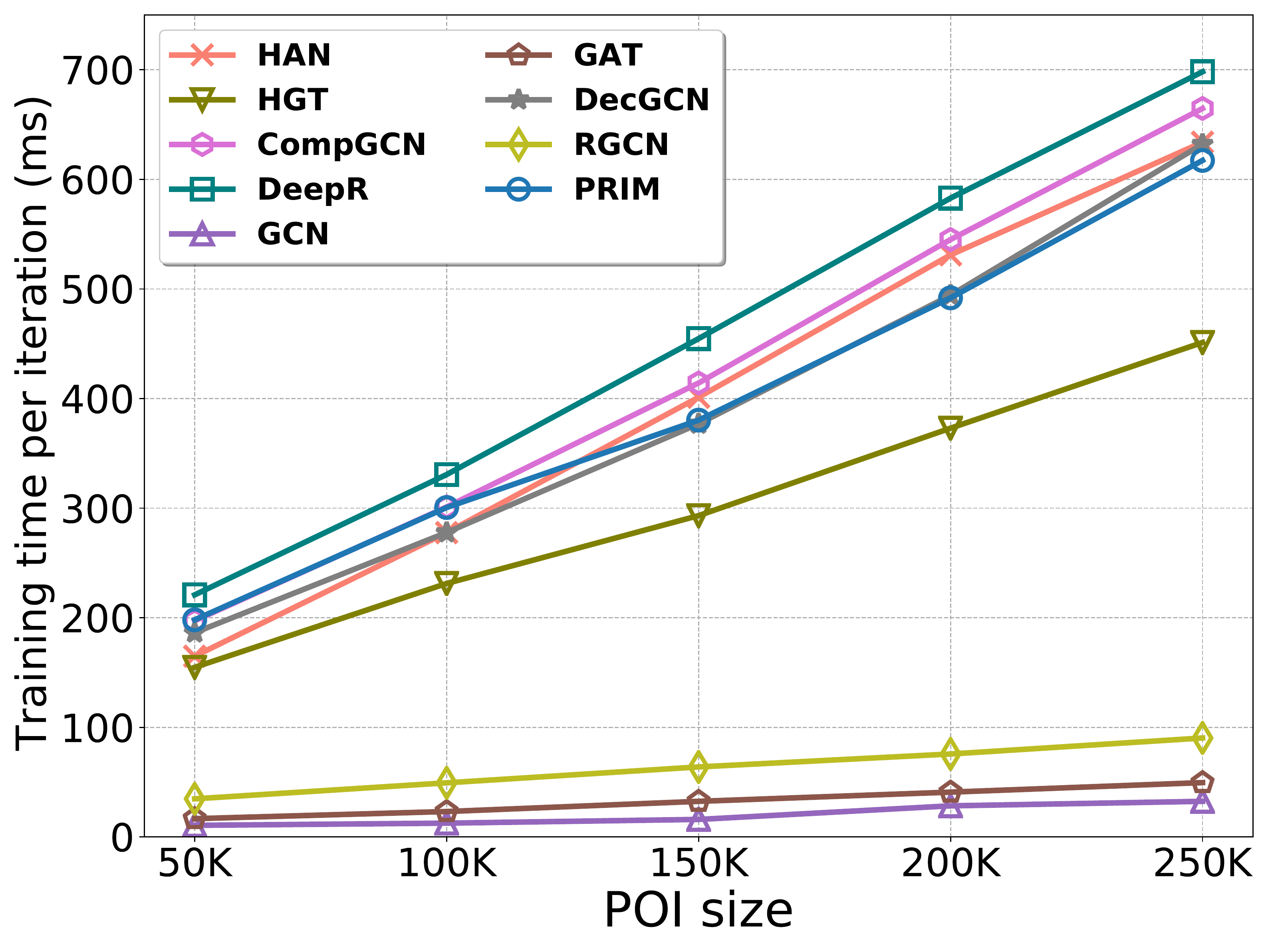}
    \vspace{-3mm}
    \caption{Training scalability}
    \label{fig:scalability}
    \vspace{-3mm}
\end{figure}

First, we evaluate the training efficiency for GNN based methods. The results are shown in Figure~\ref{fig:scalability}. It can be observed that the models that handle homogeneous graphs achieve the best efficiency.  Apart from R-GCN, all the models that can handle multiple relationships have comparable training efficiency. Furthermore, as discussed in our time complexity analysis earlier, our proposed method grows linearly against the input size in terms of the training time. The results also show that our model achieves a great trade-off between effectiveness and efficiency in training phrase.

Next, we evaluate the prediction efficiency. Given POI pairs as queries, we first obtain the embeddings for these POIs according to Section 4.2-4.4. After that, we only need to index the generated embeddings and apply scoring functions in Equation (11) \& (12) to get predictions. In this case, the prediction time does not change with the increase of POI size for all the GNN based methods. Based on such a procedure, we first generate the POI embeddings and then conduct 10K queries to test the prediction efficiency. The experiments show that the average prediction time per query is 1.57ms for our proposed method. In addition, when distance-specific hyperplane projection is not performed (Equation (11)), which is the case for the other GNN based methods, the average query time reduces to 0.61ms. In practice, both cases can satisfy the requirement in daily company services.

In summary, the training and prediction efficiency results demonstrate that our proposed method can scale to large-scale datasets and can be applied in practical use.

\begin{figure}[htbp]
\centering
\begin{subfigure}{\linewidth}
    \centering
    \includegraphics[width=\linewidth]{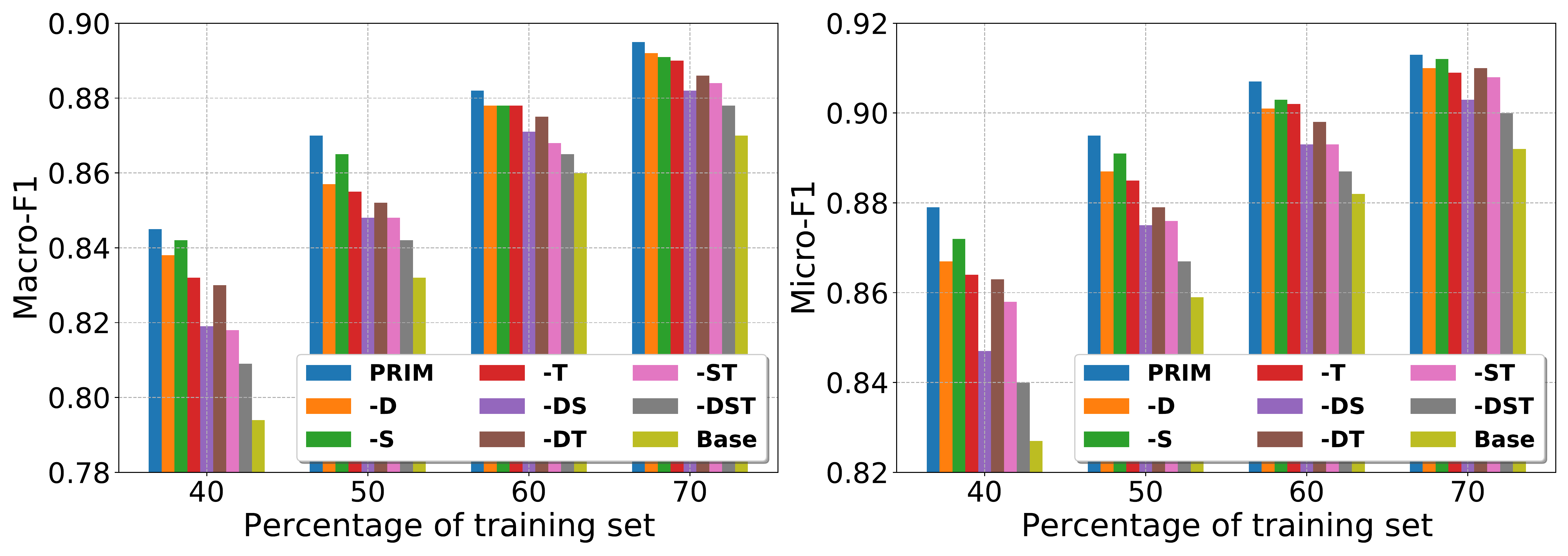}
    \label{fig:ablation_BJ}
    \vspace{-4mm}
\end{subfigure}

\begin{minipage}[t]{.48\linewidth}
\centering
\subcaption{Macro-F1 on BJ}\label{BJ_MacroF1}
\end{minipage}
\begin{minipage}[t]{.48\linewidth}
\centering
\subcaption{Micro-F1 on BJ}\label{BJ_MicroF1}
\end{minipage}
\vspace{-1mm}

\begin{subfigure}{\linewidth}
    \centering
    \includegraphics[width=\linewidth]{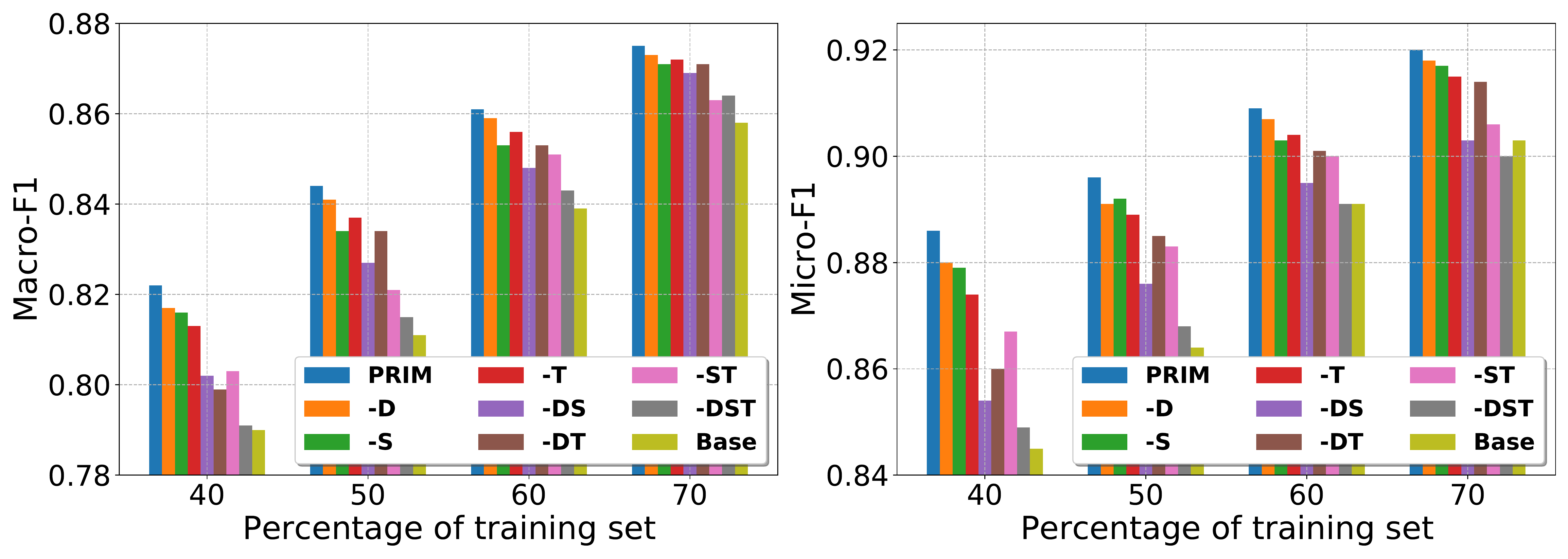}
    \label{fig:ablation_SH}
    \vspace{-4mm}
\end{subfigure}

\begin{minipage}[t]{.48\linewidth}
\centering
\subcaption{Macro-F1 on SH}\label{SH_MacroF1}
\end{minipage}
\begin{minipage}[t]{.48\linewidth}
\centering
\subcaption{Micro-F1 on SH}\label{SH_MicroF1}
\end{minipage}
\vspace{-2mm}

\caption{Results of ablation study}
\label{fig:ablation}
\vspace{-5mm}
\end{figure}

\vspace{-2ex}
\subsection{Ablation Study}
We conduct an ablation study by removing different components of \textsf{PRIM} to demonstrate their contributions to the performance. Specifically, we compare with the following model variants:  1)\textbf{-T}: the category taxonomy constraint is removed and we learn each category embeddings independently (Section~\ref{subsec:taxonomy}); 2)\textbf{-S}: the spatial context is removed (Section~\ref{subsec:spatial_context}); 3)\textbf{-D}: the distance-specific hyperplane projection (Equation~\ref{hyper}) is removed (Section~\ref{subsec:score_func}).    We also combine these three variants to test the model with two or more components being removed, i.e., \textbf{-DS}, \textbf{-DT}, \textbf{-ST} and \textbf{-DST}. In addition, we present the results of the best baseline model for comparison, denoted by \textbf{Base}.

Figure~\ref{fig:ablation} presents the results of all variants of our model, and we have the following observations: (1) The removal of different components leads to degradation of model performance, which demonstrates the effectiveness of each component. Moreover, with more components removed, we can see a larger decrease in model performance. This shows that different components capture different aspects of characteristics in POIs which are complementary to enhance the performance. (2) The performance gap between model variants becomes larger when dealing with less training data. It reveals that context information extracted from different components would be important to alleviate the issue when the data is relatively small. (3) The model with all the three components removed (equivalent to WRGNN) still outperforms the best baseline in most cases, which validates the superiority of WRGNN over baseline models.

\subsection{Model Analysis}
\subsubsection{Analysis on sparse cases}
To achieve practical usage in applications, \textsf{PRIM} is expected to be robust to all cases. That is to say, the model should deal with POIs effectively even if they have very few relationships (i.e., sparsity). For this purpose, we construct the test set for POIs with fewer than 3 relationships in the training data to check their performance. We only report the results for 4 best performing baselines, since others show consistently worse results. 

\begin{figure}[htbp]
\centering
\begin{subfigure}{\linewidth}
    \centering
    \includegraphics[width=\linewidth]{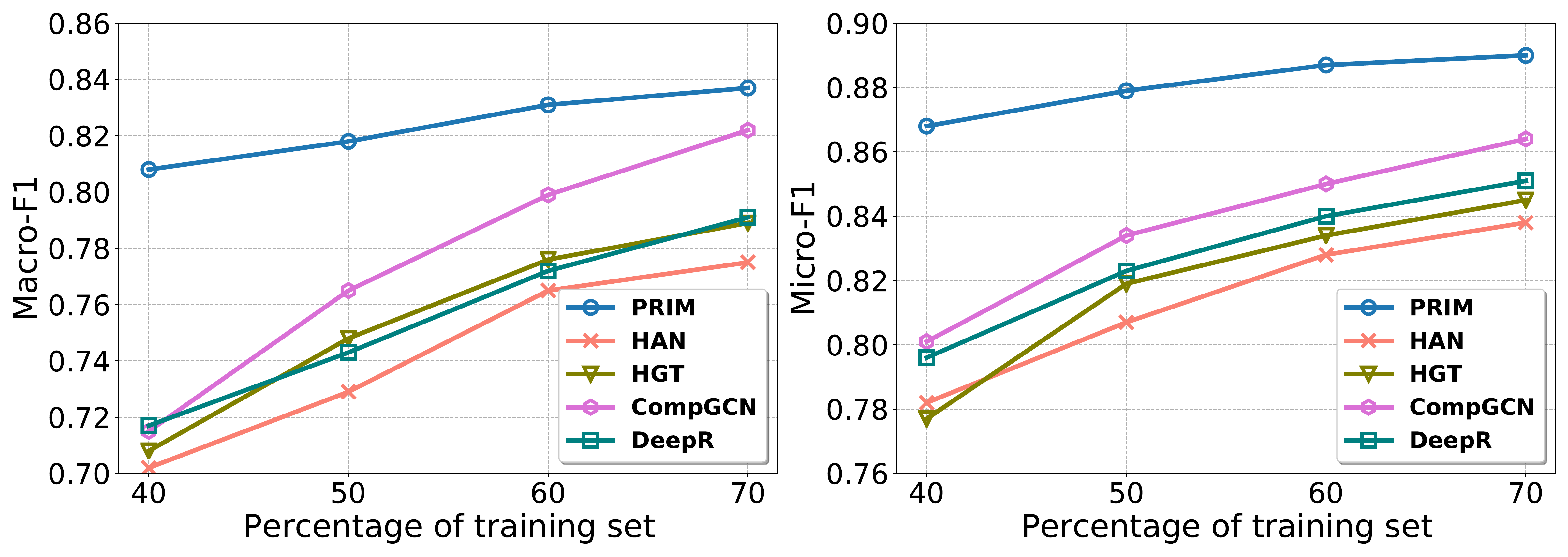}
    \label{fig:sparse_BJ}
    \vspace{-4mm}
\end{subfigure}

\begin{minipage}[t]{.48\linewidth}
\centering
\subcaption{Macro-F1 on BJ}
\end{minipage}
\begin{minipage}[t]{.48\linewidth}
\centering
\subcaption{Micro-F1 on BJ}
\end{minipage}
\vspace{-1mm}

\begin{subfigure}{\linewidth}
    \centering
    \includegraphics[width=\linewidth]{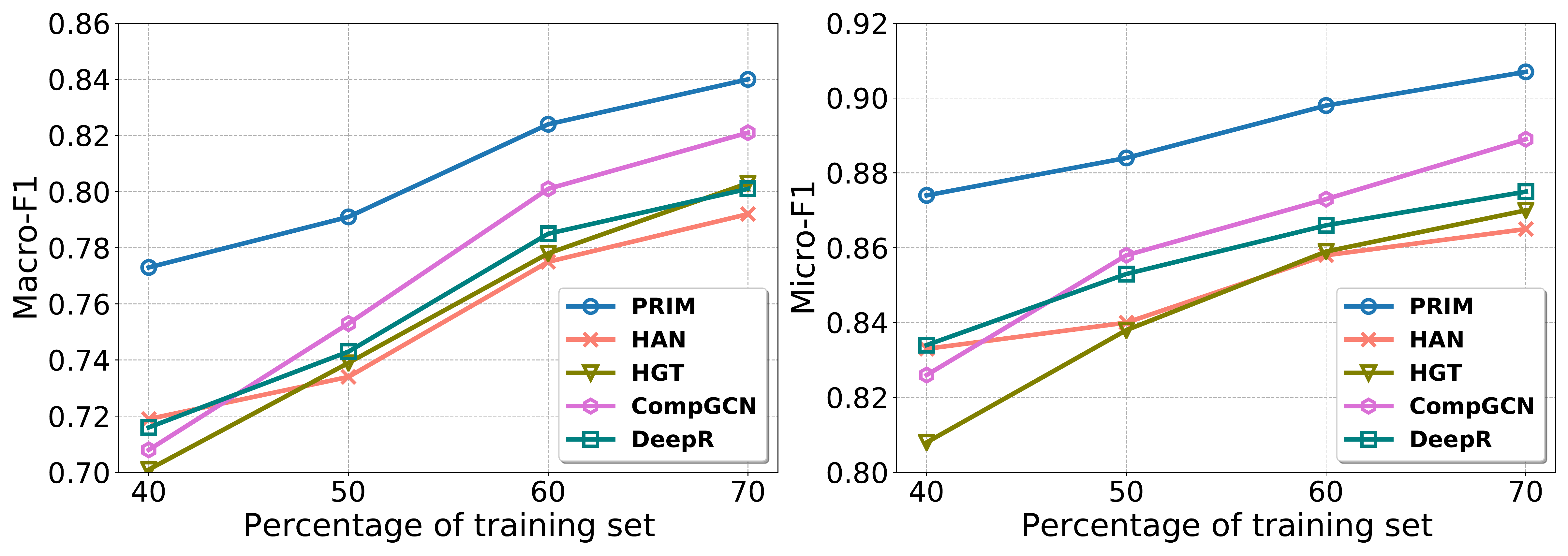}
    \label{fig:sparse_SH}
    \vspace{-4mm}
\end{subfigure}

\begin{minipage}[t]{.48\linewidth}
\centering
\subcaption{Macro-F1 on SH}
\end{minipage}
\begin{minipage}[t]{.48\linewidth}
\centering
\subcaption{Micro-F1 on SH}
\end{minipage}
\vspace{-2mm}

\caption{Results on sparse cases}
\label{fig:sparse}
\vspace{-3mm}
\end{figure}

The results of sparse POI cases are reported in Figure~\ref{fig:sparse}. We can observe from the results that our model outperforms all the baselines on both metrics. It demonstrates that our model is more capable of handling sparse cases. Furthermore, the performance of the baselines degrades more significantly than our model. 
For example, in the Shanghai dataset, the Macro-F1 decreases by 5.1\% on average for our model, and by 8.0\%, 8.4\%, 6.1\% and 7.3\% respectively for HAN, HGT, CompGCN and DeepR.
These results show the superiority of \textsf{PRIM} for exploiting the context information specific to POIs to compensate for the lack of relationships.

\subsubsection{Analysis on unseen cases}
Apart from model robustness on sparse cases, it is also important that the model should be effective in inductive settings. In other words, the model should generalize well on POIs which are not seen during training. Such a capability is desired in real scenarios since the model can be directly applied to newly arrived POIs without being frequently updated. To compare the model performance in inductive settings, we randomly hide 20\% of the POIs and remove the relation edges connecting to them in each dataset to serve as unseen cases. Then we train all the models with the remaining POIs and the same parameter settings. Same as sparse cases, we only report the results for 4 best performing baselines.

\begin{table}[htbp]
\centering
\caption{Results on unseen cases}
\vspace{-3mm}
\resizebox{0.75\linewidth}{!}{
\begin{tabular}{|c|c|c|c|c|}
\hline
Dataset & \multicolumn{2}{c|}{BJ} & \multicolumn{2}{c|}{SH} \\ \hline
        & Macro-F1   & Micro-F1   & Macro-F1   & Micro-F1   \\ \hline
HAN     & 0.844      & 0.875      & 0.794      & 0.860      \\ \hline
HGT     & 0.837      & 0.864      & 0.793      & 0.852      \\ \hline
CompGCN & 0.841      & 0.872      & 0.790      & 0.848      \\ \hline
DeepR   & 0.815      & 0.852      & 0.764      & 0.836      \\ \hline
PRIM    & \textbf{0.880}      & \textbf{0.905}      & \textbf{0.814}      & \textbf{0.885}      \\ \hline
\end{tabular}}
\label{tab:inductive}
\end{table}
\vspace{-2mm}


\begin{table}[htbp]
\centering
\caption{Model performance on different areas}
\vspace{-3mm}
\resizebox{0.9\linewidth}{!}{
\begin{tabular}{|c|c|c|c|c|c|}
\hline
\multirow{2}{*}{Metric}   & \multirow{2}{*}{Train\%} & \multicolumn{4}{c|}{Area}                     \\ \cline{3-6} 
                          &                                & BJ core area & BJ suburb & BJ overall & SH    \\ \hline
\multirow{4}{*}{Macro-F1} & 40\%                           & 0.846        & 0.844     & 0.845      & 0.722/0.822 \\ 
                          & 50\%                           & 0.871        & 0.869     & 0.870      & 0.718/0.844 \\
                          & 60\%                           & 0.881        & 0.883     & 0.882      & 0.729/0.861 \\ 
                          & 70\%                           & 0.896        & 0.894     & 0.895      & 0.741/0.875 \\ \hline
\multirow{4}{*}{Micro-F1} & 40\%                           & 0.874        & 0.890     & 0.879      & 0.797/0.886 \\ 
                          & 50\%                           & 0.890        & 0.906     & 0.895      & 0.813/0.896 \\ 
                          & 60\%                           & 0.903        & 0.915     & 0.907      & 0.801/0.909 \\ 
                          & 70\%                           & 0.911        & 0.918     & 0.913      & 0.823/0.920 \\ \hline
\end{tabular}}
\label{tab:region}
\end{table}
\vspace{-3mm}

The results of unseen POI cases are listed in Table~\ref{tab:inductive}. We observe that all the compared models achieve relatively good results. This is because they are all based on GNN models, which have been shown to handle inductive settings well~\cite{GraphSAGE}. Specifically, DeepR performs the worst among all the baselines, while HAN, HGT and CompGCN have comparable performance. In addition, ~\textsf{PRIM} outperforms all the compared methods, which verifies the effectiveness of our model on unseen POIs.   

\subsubsection{Analysis on different regions}
To gain a better understanding of our model's performance on different regions, we split Beijing into core area and suburb. The core area is less than 15\% of the area in Beijing with over 53\% POIs, which is much denser than the suburb. We report the model's performance for these two regions in Table~\ref{tab:region}. The results show that the performance gap is very small between the core area and the suburb, which demonstrates the robustness of our model on regions with different sparsity.

We further investigate the performance on regions with more complex differences by directly applying the model learned from Beijing on Shanghai. The results are presented in Table~\ref{tab:region}. In the column of Shanghai dataset, the first and second values indicate the performance of model trained on Beijing and model trained on Shanghai itself respectively. We can observe larger performance drop compared to regions within a city. However, the overall performance is still good where most Micro-F1 metrics are over 0.8.

\begin{figure}[htbp]
\centering
\begin{subfigure}{\linewidth}
    \centering
    \includegraphics[width=\linewidth]{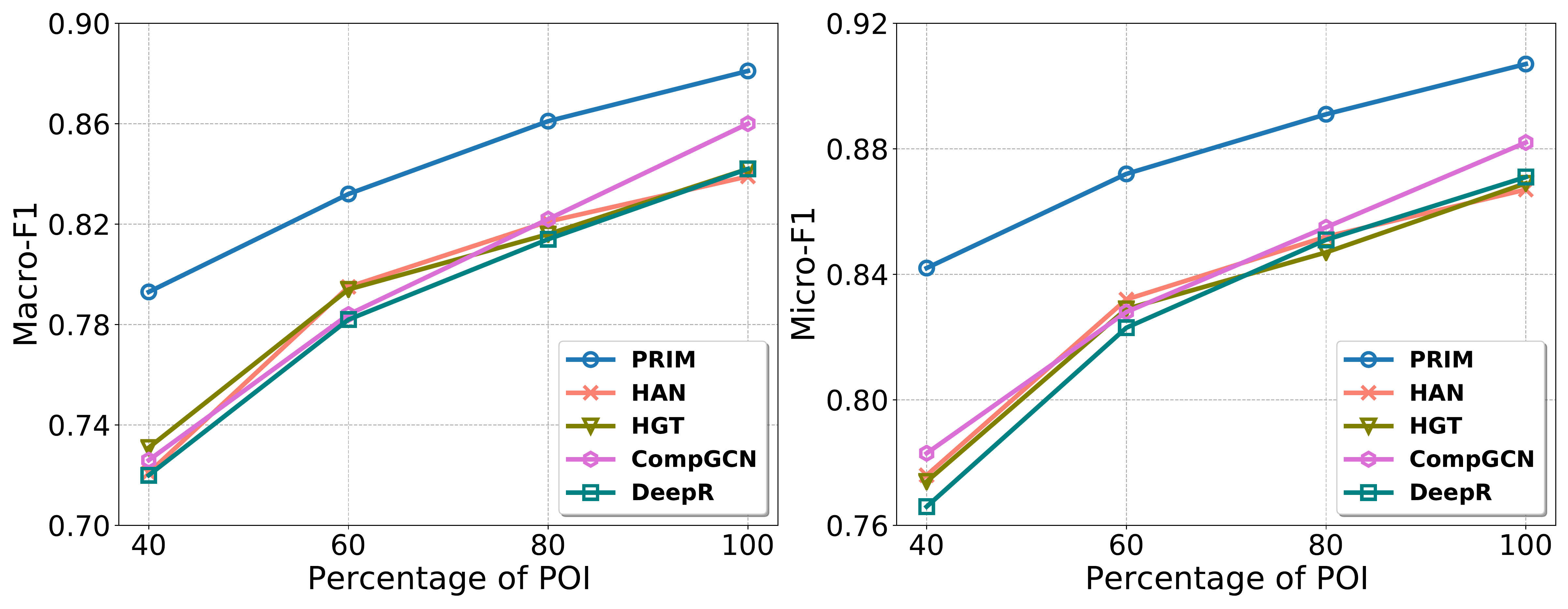}
    \label{fig:different_scale}
    \vspace{-4mm}
\end{subfigure}

\begin{minipage}[t]{.48\linewidth}
\centering
\subcaption{Macro-F1}
\end{minipage}
\begin{minipage}[t]{.48\linewidth}
\centering
\subcaption{Micro-F1}
\end{minipage}
\vspace{-2mm}

\caption{Results on datasets with different characteristics}
\label{fig:different_scale}
\vspace{-3mm}
\end{figure}

\subsubsection{Analysis on datasets with different characteristics}

We conduct experiments to evaluate the model performance on datasets with different characteristics. To this end, we propose to generate datasets by sampling subsets of the Beijing dataset. Specifically, we randomly select 40\%/60\%/80\% POIs in Beijing and only keep the edges among the selected POIs to serve as datasets with different scale/density/spatial distances. As a result, sparser datasets have lower density and larger spatial distances. Then we split the relational edges into 60\%/20\%/20\% as train/validation/test data. The experimental results are shown in Figure~\ref{fig:different_scale}. We can observe that $\textsf{PRIM}$ outperforms all the baselines, thus validating the superiority of our proposed model in handling datasets with different characteristics.

\section{Conclusion}

In this paper, we proposed a new model, PRIM, to solve the POI relationship inference problem with multiple relation types. It is featured with several new components, including 
the weighted relational graph neural networks, incorporating structural constraints from category taxonomy, a self-attentive spatial context extractor, and self-attentive spatial context extractor.
%
Extensive experiments on two real-world datasets have demonstrated the superiority of the PRIM model over state-of-the-art baselines.

\bibliographystyle{ACM-Reference-Format}
\bibliography{main}

\end{document}